# Quadratic Unconstrained Binary Optimization Problem Preprocessing: Theory and Empirical Analysis

Mark Lewis

Missouri Western State University, Saint Joseph, Missouri 64507, USA

Fred Glover

School of Engineering & Science, University of Colorado, Boulder, Colorado 80309, USA

**Abstract**.  The Quadratic Unconstrained Binary Optimization problem (QUBO) has become a unifying model for representing a wide range of combinatorial optimization problems, and for linking a variety of disciplines that face these problems. A new class of quantum annealing computer that maps QUBO onto a physical qubit network structure with specific size and edge density restrictions is generating a growing interest in ways to transform the underlying QUBO structure into an equivalent graph having fewer nodes and edges.  In this paper we present rules for reducing the size of the QUBO matrix by identifying variables whose value at optimality can be predetermined.   We verify that the reductions improve both solution quality and  time to solution and, in the case of metaheuristic methods where optimal solutions cannot be guaranteed, the quality of solutions obtained within reasonable time limits.

We discuss the general QUBO structural characteristics that can take advantage of these reduction techniques and perform careful experimental design and analysis to identify and quantify the specific characteristics most affecting reduction.  The rules make it possible to dramatically improve solution times on a new set of problems using both the exact Cplex solver and a tabu search metaheuristic.

## 1. INTRODUCTION

Given is the graph G = [N, E] with node set N = { 1, 2, …, $i$, … n } and edge set E = {($i,j$): $i, j \in$ N }.  Denoting the weight of edge ($i, j$) by $c_{ij}$, we define the Quadratic Unconstrained Binary Optimization Problem (QUBO) as:

Maximize: $\sum_{i \in N} c_{ii} x_i + \sum_{(i,j) \in E} c_{ij} x_i x_j$    subject to $x_i = \{0,1\}$ where $i \in$ N          (1)

The equivalent compact definition with the coefficients of (1) represented as a $Q$ matrix is:

$$\text{Max } x^t Q x: \ x \in \{0, 1\}^n$$

where $Q$ is an $n$-by-$n$ square symmetric matrix of coefficients.

## 2. LITERATURE

QUBO has been extensively studied [12] and is used to model and solve numerous categories of optimization problems including important instances of network flows, scheduling, max-cut, max-clique, vertex cover and other graph and management science problems, integrating them into a unified modeling framework [11]. Many NP-complete problems such as graph and number partitioning, covering and set packing, satisfiability, matching, spanning tree as well as others can be converted into the Ising form as shown in [14]. Ising problems replace $x \in \{0, 1\}^n$ by $x \in \{-1, 1\}^n$ and can be put in the form of (1) by defining $x_j' = (x_j + 1)/2$ and then redefining $x_j$ to be $x_j'$.[1] Ising problems are often solved with annealing approaches in order to find a lowest energy state.

Although QUBO problems are NP-complete, good solutions to large problems can be found using modern metaheuristics [8]. In addition, a new type of quantum computer based on quantum annealing with an integrated physical network structure of qubits known as a Chimera graph has also been demonstrated to very quickly find good solutions to QUBO [4]. The Chimera structure is a connected network of qubits with groups of densely connected nodes sparsely connected to other groups of densely connected nodes, having a structure similar to that of social network visualizations or to a collection of densely connected cities sparsely linked to other cities via fiber optic backbones (see Figure 1). Transforming a given problem graph by mapping it onto all or part of the Chimera hardware graph requires minor-embedding and is described in [7].

A set of rules for reducing multi-commodity networks based on the structure of the network [9] has generated interest in investigating whether similar rules could be found for QUBO. For certain classes of very structured problems such as vertex cover, max-cut and max-clique, the work of [6] shows that complete reduction can be achieved via computation of the roof duals of the associated capacitated implication network in association with rules involving first and second order derivatives. Similarly, maximum flow and multi-commodity flow networks can be used to help determine QUBO optimal variable assignments and lower bounds [1,19]. In comparison, we present and test four basic rules, iteratively applying them to reduce the size of the $Q$ matrix until no further reductions are possible. We also explore transformations to reduce a node's edge density (with application to hardware graphs such as the Chimera) and discuss applications to sensitivity analysis.

---

[1] This adds a constant to (1), which is irrelevant for optimization.

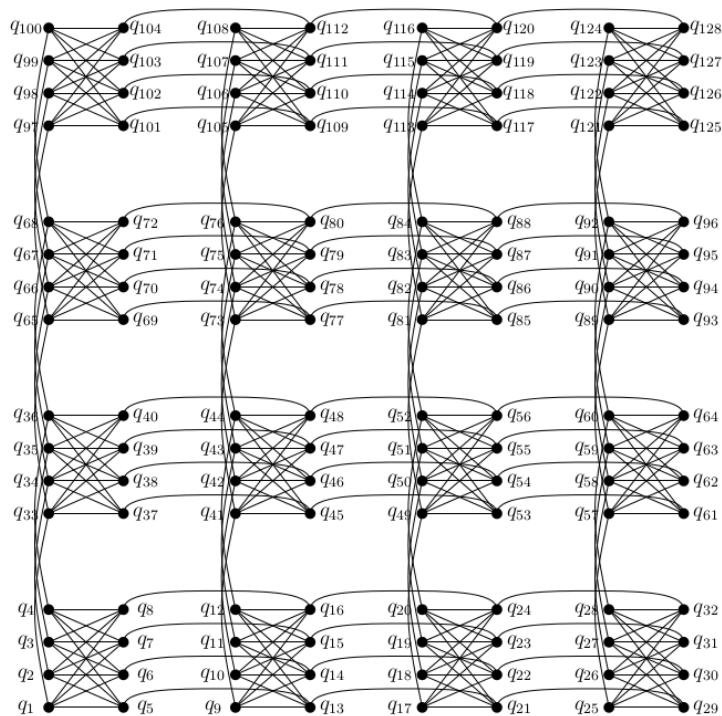

Figure1. Example Chimera Network Structure

Benchmark QUBO problems are often highly structured, or have uniform coefficient distributions, or are dense, or are randomly connected [2,16]. Classic problems with wide application such as the maximum cut problem are highly structured, e.g., all quadratic coefficients are negative and all linear coefficients are positive, or quadratic coefficients are -1s and linear coefficients are positive sums of quadratic coefficients. The rules presented here for predetermining the optimal assignment of variables are applicable to any QUBO, but in this paper we apply them to $Q$ matrices having structural characteristics associated with real-world graphs (sometimes called complex networks [10]) grounded on assumptions from experimental design [18], namely that there are random elements with a small percentage of variables having strong effects. This is known as the "*sparsity of variable effects*" [15] and states that, in general, when many factors are examined for their effect on a performance parameter (i.e., objective function), a relatively small percent have large effects. The Pareto Principle is similar, stating that a small percent of causes account for the majority of effects.

Thus, we investigate problems in which $Q$ is connected, generally sparse but with some densely connected nodes, mostly uniform in distribution but containing a small percent of linear and quadratic elements falling outside the limits of the majority uniformly distributed elements. A histogram of a typical distribution based on 1000 nodes and 5000 edges is shown as the columns in Figure 2. The original *Q* has most elements uniformly distributed between -10 and 10, with a small number of outliers with magnitudes between 25 and 250. The reduced *Q* distribution (dashed line) has removed many of these outliers to yield a *Q* with a reduced number of nodes and edges. This is the first time problems of this type have been studied in the literature. The *Q* generator code and the experimentally designed test set and network generator are available at the first author's website.

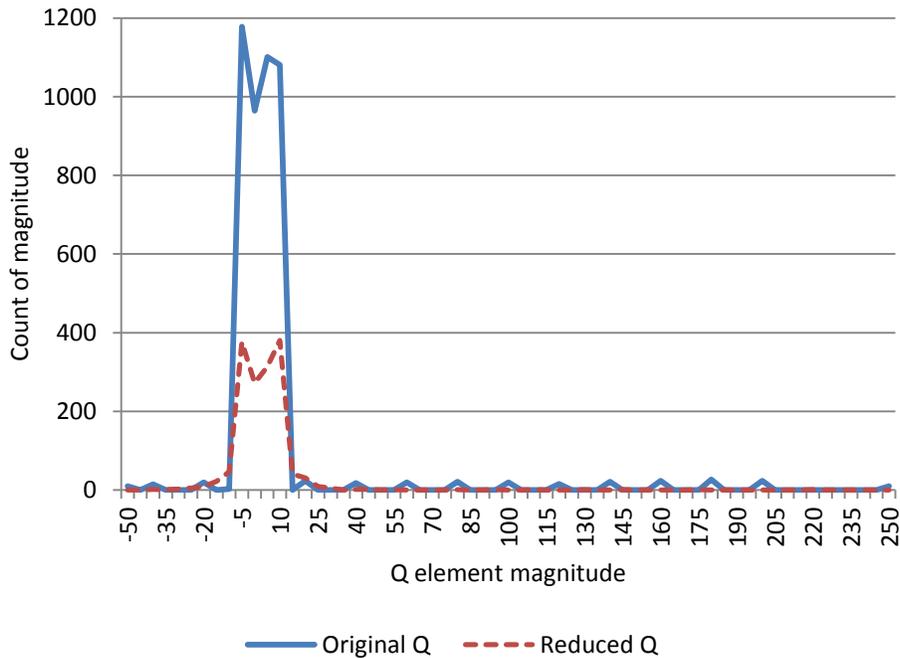

Figure 2. Distribution of *Q* with Outliers Before and After Reduction.

The remainder of this paper is organized as follows. Section 3 presents the rules for reducing *Q*. Section 4 discusses network transformations when nodes have an upper limit on the number of incident edges. Modifications of the rules to define the range over which coefficients can change while enabling a transformation to remain valid are presented in Section 4.2. Section 5 provides the pseudocode used to implement the rules which we have embodied in our preprocessor, named QPro, and Section 6 presents the experimental design factors, tests run parameters and analyzes the test results based on Cplex and a path relinking metaheuristic.

## 3.  RULES FOR REDUCING *Q* TO SHRINK QUBO

The major rules for reducing *Q* are provided below.  After stating the rules, we provide an efficient implementation followed by testing.  Future research will investigate further enhancements and implementation trade-offs.  We employ the following notation.

Let $N_i^+ = \{j \in N: c_{ij} > 0, i{\neq}j\}$, $C_i^+ = \sum(c_{ij}: j \in N_i^+)$, $N_i^- = \{j \in N: c_{ij} < 0, i{\neq}j\}$ and $C_i^- = \sum(c_{ij}: j \in N_i^-)$. By convention, a summation over an empty set equals 0. Hence $C_i^+ = 0$ or $C_i^- = 0$, respectively, if $N_i^+$ or $N_i^-$ is empty.

***Rule 1***: (For $c_{ii} \geq 0$.) If $c_{ii} + C_i^- \geq 0$, then $x_i = 1$ in an optimal QUBO solution.

Rule 1 is based on the simple observation that if $x_i = 1$, the least possible contribution to the objective function is created by setting $x_j = 0$ for all $j \in N_i^+$ and $x_j = 1$ for all $j \in N_i^-$, yielding $c_{ii} + C_i^-$. If this quantity is $\geq 0$ then evidently there is no loss in setting $x_i = 1$ and the conclusion of Rule 1 holds. (The condition $c_{ii} \geq 0$ is implied by $c_{ii} + C_i^- \geq 0$.)  When Rule 1 is satisfied  $c_{ii}$ is added to the objective function, the $c_{ij}$ coefficients are added to the corresponding diagonal coefficients $c_{jj}$ and row *i* and column *i* are removed from the *Q* matrix.

***Rule 2***: (For $c_{ii} \leq 0$.) If $c_{ii} + C_i^+ \leq 0$, then $x_i = 0$ in an optimal QUBO solution.

Similarly, Rule 2 is based on the observation that if $x_i = 1$, the greatest possible contribution to the objective function occurs by setting $x_j = 1$ for all $j \in N_i^+$ and $x_j = 0$ for all $j \in N_i^-$, yielding $c_{ii} + C_i^+$. If this quantity is $\leq 0$ then there can be no gain by setting $x_i = 1$ and hence the conclusion of Rule 2 holds. (The condition $c_{ii} \leq 0$ is implied by $c_{ii} + C_i^+ \leq 0$.)  When Rule 2 is satisfied row and column *i* can be removed from the *Q* matrix to create a reduced Q.   There is no adjustment to the objective function.

We observe in the extreme case, where $c_{ii} = 0$ yields $x_i = 0$ in Rule 1 or $x_i = 0$ in Rule 2, then $N_i^-$ or $N_i^+$ is empty, respectively.  We assume the indexes *i* and *h* in all subsequent rules are distinct.

***Rule 3***: Assume Rule 1 does not yield either $x_i = 1$ or $x_h = 1$. If $c_{ih} > 0$ ($h \in N_i^+$ and $i \in N_h^+$) and if $c_{ii} + c_{hh} + c_{ih} + C_i^- + C_h^- \geq 0$, then $x_i = x_h = 1$ in an optimal QUBO solution.

The justification of Rule 3 is as follows. If Rule 1 does not yield $x_i = 1$ or $x_h = 1$, then $c_{ii} + C_i^-$ and $c_{hh} + C_h^-$ are both negative, and the condition $c_{ii} + c_{hh} + c_{ih} + C_i^- + C_h^- \geq 0$ implies $c_{ih} > 0$ and consequently $h \notin N_i^-$ and $i \notin N_h^-$.  As previously noted, the least possible contribution to the objective function when $x_i = 1$ results by setting $x_j = 0$ for all $j \in N_i^+$ and $x_j = 1$ for all $j \in N_i^-$, and similarly the least possible contribution to the objective function when $x_i = 1$ results by setting $x_j$

= 0 for all j $\in N_h^+$ and $x_j = 1$ for all j $\in N_h^-$.  Hence we can obviously do no worse for $x_i = x_h = 1$ than to achieve the value $c_{ii} + c_{hh} + c_{ih} + C_i^- + C_h^-$ and if this value is nonnegative the objective function is not reduced.

**Rule 4**: Assume that Rule 2 does not yield either $x_i = 0$ or $x_h = 0$. If $c_{ih} > 0$ (h $\in N_i^+$ and i $\in N_h^+$) and if $c_{ii} + c_{hh} + c_{ih} + C_i^+ + C_h^+ \leq 0$, then $x_i + x_h \leq 1$ holds in an optimal QUBO solution.

The justification of Rule 4 derives from an analysis related to the arguments justifying the preceding rules.  Rule 4 does not predetermine a variable's value and was not implemented in this investigation; however future research will investigate the transformation of $c_{ih}$ to penalize the occurrence of $x_i = x_h = 1$.

**Rule 5:**  This is the trivial case when a row in the $Q$ matrix is all 0s.  In this case neither $x_i = 0$ nor $x_i = 1$ has an objective value effect and $x_i$ can be eliminated from $Q$.  Although you would not expect to create a QUBO with this condition, it may occur during preprocessing transformations.

## 4. GRAPH EXPANSION AND SENSITIVITY ANALYSIS

### 4.1 Graph Expansion via Strongly Coupled Nodes

In practice it is possible that a node may be restricted in the number of incident edges.  This occurs in quantum annealing computers as well as in communication networks where nodes have edge capacity limitations.  In these cases, the over-capacity node moves some of its edges to additional nodes that are *strongly coupled* to it so that all have the same value at optimality.   Let *m* be the maximum allowable number of edges incident to any given node in the set of nodes N.  Let $E_i$ be the subset of node pairs in E that contain the node *i*,  $E_i = \{(k, l): k = i \ or \ l = i\}$.  So $|E_i|$ is the number of edges incident to node *i* and the restriction is $|E_i| \leq m$.

If there exist nodes in G having $|E_i| > m$, then G can be transformed to an expanded graph G* = [N*, E*] via the introduction of additional nodes *n** that are strongly coupled to those nodes having more than the allowable edges $|E_i| > m$.  Thus N* = {1, 2, …, *i*, … n, (n+1)*, (n+2)*, …, n*} contains the original nodes in N up to n, but we will rearrange the edges between nodes in N* to accommodate the additional nodes (n+1) to n*.

When mapping to a physical graph such as the Chimera graph used in quantum annealing computers [17], we assume that G* is also subject to $|E_i| \leq m$ and transformations can be continued, if necessary, until $|E_i| = m$.  The optimal solution to the QUBO problem based on the original G will be equivalent to the optimal solution based on G*.

In order to strongly couple a collection of nodes we make use of penalty functions described in [11]. Specifically, if we wish to strongly couple nodes $i$ and $j$ in G*, then we use the penalty function $M(x_i - 2 x_i x_j + x_j)$ in the objective function, where M is a large negative number in a maximization. Note that the distinction between strong coupling and our Rule 3 is that the latter forces the corresponding variables to be equal to 1 in the optimal solution while strong coupling forces them to have the same value, either 0 or 1 at optimality.

A small example is presented to illustrate the transformation of G to G* via the addition of strongly coupled nodes. Figure 3 shows the edges between nodes of a small graph G with 5 nodes. Let $m = 3$, that is, a node can have at most 3 edges. However node 1 has 4 edges, therefore the graph will be transformed by adding a node (or nodes) with penalty functions that guarantee that the optimal solution to both G and G* are equivalent. Note that there can be multiple ways to add nodes $n*$ and the edges linking the original and new nodes.

Figure 3 illustrates two transformations; the first adds a single node $x_6$ with the maximum 3 edges (strongly coupled edge is bolded). The second transformation adds two nodes $x_6$ and $x_7$ leaving an open edge on node 7 to which other nodes can be added.

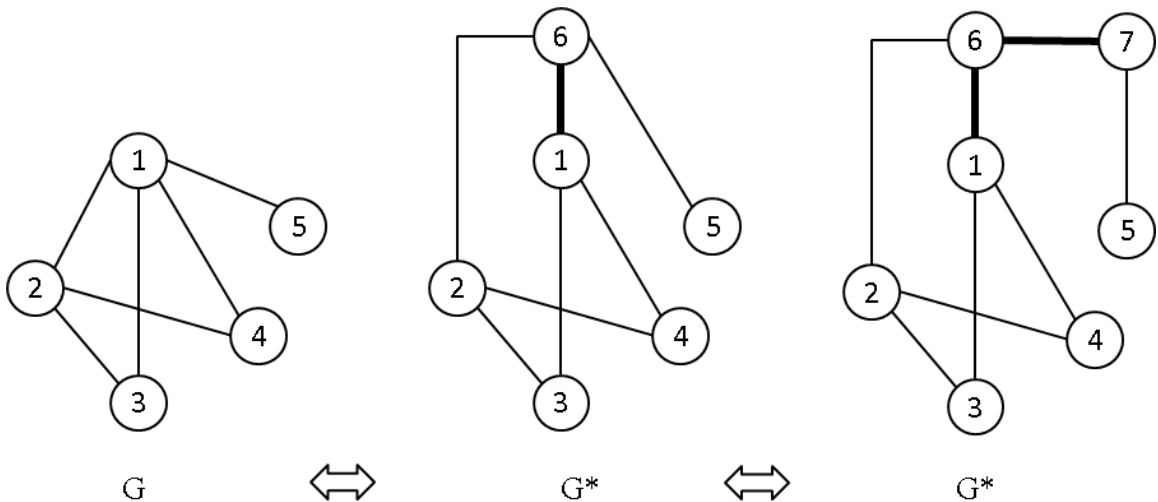

Figure 3. Mapping from G to G* when $m = 3$

In practice, we add an element $x_{n+1}$ to the $Q$ matrix that is strongly coupled to any node $i$ with the following elements modified based on the value of penalty term M. Figure 4 provides an example of the changes made when adding node 6 that is strongly coupled to node 1.

$$G^*(c_{i,i}) \;=\; G(c_{i,i}) + M$$
$$G^*(c_{n+1,n+1}) \;=\; -2M$$
$$G^*(c_{i,n+1}) = M$$

|   | 1 | 2 | 3 | 4 | 5 |
|---|---|---|---|---|---|
| 1 | 5 | 2 | 2 | 2 | 2 |
| 2 |   | 8 | 2 | 2 |   |
| 3 |   |   | 3 |   |   |
| 4 |   |   |   | -2 |   |
| 5 |   |   |   |   | -4 |

$\Longleftrightarrow$

|   | 1 | 2 | 3 | 4 | 5 | 6 |
|---|---|---|---|---|---|---|
| 1 | -45 |   | 2 | 2 |   | 100 |
| 2 |   | 8 | 2 | 2 |   |   |
| 3 |   |   | 3 |   |   |   |
| 4 |   |   |   | -2 |   |   |
| 5 |   |   |   |   | -4 |   |
| 6 |   |   |   |   |   | -50 |

Original Q                Transformed Q

Figure 4.  Small example transformation when strongly coupling nodes 1 and 6

Future work will investigate the application of Rules 1-5 in conjunction with strong coupling in order to transform a given graph to one that meets a target graph's node and edge specifications [5].

### 4.2 Use of the Rules in Sensitivity Analysis

Robust optimization [3] is concerned with the fact that most data sets have a random element and thus contain inaccuracies and should not be treated as precise.  Since models using inaccurate data can lead to suboptimal solutions, the robustness of a solution to changes in the data should be examined.  Knowing the range of values over which a variable is determined as well as the relationship between that range of values and the interacting elements is a fundamental component of robustness.

We examine Rules 1-4 to see how they are useful for analyzing the sensitivity of a determined variable to changes in elements of Q.   The rules provide the magnitude of change needed for a variable to become determined, or to stay indeterminate.

Let $\Delta c_{ij}$ denote a change in the current value of $c_{ij}$ and set $\Delta c_{ij} = 0$ to yield an alternative expression of Rule 1.  For a given $i$ where $c_{ii} \geq 0$,

$$\text{if } c_{ii} \geq \sum_{\substack{i,j \\ i \neq j}} |c_{ij} + \Delta c_{ij}| \text{ where } c_{ij} < 0 \text{ and } \Delta c_{ij} = 0, \text{ then } x_i = 1 \qquad \text{(R1a)}$$

For a given $i$ where R1 is valid, based on R1a the *allowable decrease* $\Delta c_{ij}$ to $c_{ij}$ for a given $j$ and still having $x_i = 1$ determined is

$$\Delta c_{ij} \leq \sum_{\substack{i,k \\ i \neq k}} |c_{ik}| - c_{ii} \quad \forall k = 1 \ldots n \tag{R1b}$$

Thus, the right hand side of R1b is the difference in magnitude between the linear coefficient of a row and the sum of the negative quadratic coefficients for that row. Conversely if an $x_i$ is not determined, then R1b provides the amount that either $c_{ii}$ must increase, or the amount that a $\Delta c_{ij}$ must decrease in order for $x_i$ to be set to 1 as shown in R1c. By extension the sum of the negative changes to *all* negative interactions can only decrease by the amount of the right hand side of R1b in order for $x_i$ to remain set to 1.

$$\sum_{\substack{i,j \\ i \neq j}} \Delta c_{ij} \leq \sum_{\substack{i,k \\ i \neq k}} |c_{ik}| - c_{ii} \quad \forall k = 1 \ldots n \tag{R1c}$$

Similar expressions can be developed for Rules 2 and 3. Although we did not specifically investigate sensitivity analysis and robustness in this paper, we did perform some repeated testing using a random $Q$ matrix to verify the robustness of certain results (see Section 6).

## 5. PSEUDOCODE

An implementation of rules 1-5 is outlined below and then described in more detail.

```
Inputs:  graph G of size n
Outputs: graph G* of size n*

1.  Convert_G_to_Q;  // read graph and convert to an internal Q format
2.  sum_of_positive_off_diagonal[i] = Calculate_pos_sum_in_row( i );
3.  sum_of_negative_off_diagonal[i] = Calculate_neg_sum_in_row( i );
4.  x_determined[ i ] = -1;  // indicates whether variable i = 0,1, unknown
5.  number_determined = -1;
6.  While number_determined <> 0
7.      number_determined = Determine_x; // applies Rules 1-5
8.      Q = Reduce_Q;  // reduce the size of Q and adjust cᵢᵢ and cᵢⱼ
9.      Adjust_objective_function_value;
10. Save_reduced_G*;
```

Step 1 is provided to address the various formats for describing nodes and edges in a file and various methods for working with the $Q$ matrix, e.g., input is provided as a full matrix or in row-col-value format and stored in memory as a full or upper triangular matrix, hash table, or linked list. Step 2 calculates $\sum_{\substack{i,j \\ i \neq j}} c_{ij}$ where $c_{ij} > 0$ for each $i \in \mathrm{N}$ and Step 3 calculates $\sum_{\substack{i,j \\ i \neq j}} c_{ij}$ where $c_{ij} < 0$ for each $i \in \mathrm{N}$. These sums are dynamic and are updated in Step 8. Step 4 initializes an array recording whether a variable has been set to 0 or 1 or has not been determined (set to -1). Step 7 implements the Rules 1-5 and maintains the array of determined variables. Step 8 reduces $Q$ based on the results of Step 7 and updates the sums calculated in Steps 2 and 3. Any variables determined to equal 1 require that the objective function be adjusted by a constant in Step 9. As $Q$ is transformed, new determinable variables can be discovered, which continues until none are determined (Steps $6 - 9$).

## 6. TESTING

As noted in [6] "the border separating successful from unsuccessful preprocessing cases is very thin." To gain an understanding of what separates successful from unsuccessful preprocessing, an experimental design approach was used to identify the main $Q$ characteristics affecting QPro efficacy. Six $Q$ factors, or characteristics, were considered for their effect on three outputs of interest: percent $Q$ reduction, objective value quality and time to best solution. The factors and their settings used in the experimental design are described in Table 1. We created a $2^{6-2}$ fractional factorial design resulting in 16 tests for each of the 3 problem sizes and 2 problem densities, creating a total of 96 test problems with detailed results provided in the Appendices.

Table 1. $Q$ Factors and their Low / High Settings

| Factor ID | Description | Low | High |
|-----------|-------------|-----|------|
| 1 | -Upper Bound $< c_{ij} <$ Upper Bound | 10 | 100 |
| 2 | Linear Multipliers | 5 | 10 |
| 3 | Quadratic Multipliers | 10 | 20 |
| 4 | % Quadratic Multiplied | 5 | 15 |
| 5 | % Linear Multiplied | 10 | 20 |
| 6 | % non-zero Linear elements | 5 | 25 |

The six factors in the table affect the magnitude, distribution and density of the $c_{ij}$ in $Q$. During problem generation these quantities are set to one of two values. A description of the six factors and their effect follows. The first factor sets the range of the uniform random number generator; for example a setting of 10 indicates that random coefficients $c_{ij}$ are uniformly distributed between -10 and +10. The second factor is multiplied times the number generated within the bounds of factor 1 according to the probability percent of factor 5. For example, setting factors 2 and 6 to their low settings means 5% of the linear elements are multiplied by 5 when generating the $Q$ matrix, where factor 6 indicates what percentage of the $Q$ matrix will have linear elements. Factor 3 is similar to factor 2 except it is used for quadratic elements and factor 4 determines the percentage of quadratic elements that will become outliers. Thus the majority of $Q$ elements are drawn from a uniform distribution but with a percentage of them moved outside the limits of uniformity.

For the three problem sizes we defined two different edge densities with characteristics provided in Table 2. A description of the 16 test runs based on the parameter settings in Table 1 are listed in Table 3. The column headings refer to the parameter descriptions in Table 1. The problem generator creates edges similar to the $Q$ coefficients, in that they are uniformly distributed except that 1% of the nodes are densely connected. While the average densities may seem small, they represent up to 50 edges per node (P6), implying a binary decision quantifiably interacting with 50 other decisions. All problems generated are connected graphs, but it is apparent that during preprocessing the graph could become disconnected, which would create multiple independently solvable smaller problems and future research will explore how to best leverage this fact.

Table 2. Problem Characteristics

| Problem ID | $Q$ size | Edges | Density % |
|---|---|---|---|
| P1 | 1000 | 5000 | 1 |
| P2 | 1000 | 10000 | 2 |
| P3 | 5000 | 25000 | 0.2 |
| P4 | 5000 | 50000 | 0.4 |
| P5 | 10000 | 100000 | 0.2 |
| P6 | 10000 | 500000 | 1 |

Table 3. Experimental Design Factors for 16 Tests

| ID | Upper limit | Linear Factor | Quadratic Multiplier | % large Quadratic | % large Linear | % non-zero Linear |
|---|---|---|---|---|---|---|
| 1 | 10 | 10 | 20 | 5% | 10% | 25% |
| 2 | 100 | 10 | 20 | 15% | 20% | 25% |
| 3 | 10 | 5 | 20 | 15% | 10% | 5% |
| 4 | 100 | 5 | 20 | 5% | 20% | 5% |
| 5 | 10 | 10 | 10 | 5% | 20% | 5% |
| 6 | 100 | 10 | 10 | 15% | 10% | 5% |
| 7 | 10 | 5 | 10 | 15% | 20% | 25% |
| 8 | 100 | 5 | 10 | 5% | 10% | 25% |
| 9 | 100 | 5 | 10 | 15% | 20% | 5% |
| 10 | 10 | 5 | 10 | 5% | 10% | 5% |
| 11 | 100 | 10 | 10 | 5% | 20% | 25% |
| 12 | 10 | 10 | 10 | 15% | 10% | 25% |
| 13 | 100 | 5 | 20 | 15% | 10% | 25% |
| 14 | 10 | 5 | 20 | 5% | 20% | 25% |
| 15 | 100 | 10 | 20 | 5% | 10% | 5% |
| 16 | 10 | 10 | 20 | 15% | 20% | 5% |

*6.1  Test Results Using Cplex*

The 96 problems were first solved by default Cplex (with the quadratic-to-linear parameter turned off so that the problems were not linearized) and compared to using QPro followed by solution of the reduced problem using Cplex.   Default Cplex presolve was used for both approaches (except the quadratic-to-linear parameter was set to zero) and the average percent reductions found by QPro alone and by Cplex are summarized in Table 4 with detailed test results available in Appendix A.  Tests were performed using 64 bit Windows 7 on an 8-core i7 3.4 GHz processor with 12 GB RAM.

Table 4 summarizes the data from the 16 test runs for the 3 problem sizes and 2 densities and provides the time taken by QPro to perform the reduction (first column), the total QPro + Cplex time and the percent reduction of $Q$.  For example, the first row shows that the average QPro time was 0.01 seconds, with total QPro + Cplex time being 4 seconds and the average reduction to $Q$ was 36%.   The average speed up in time for QPro + Cplex was 7x, QPro's reductions were on average 3x greater and the average objective difference was zero.    Table 4 shows that QPro reduced all $Q$ matrices on average by 30% while Cplex's percent reduction was about 5%. The table also shows that QPro + Cplex was faster to obtain the same, or better, solutions.  The objective differences reported in Table 4 for the 10000 variable 100000 edge problems are noticeably higher because of two tests (84 and 95 in Appendix A) where QPro+ Cplex found much better answers to problems with large objectives.  However, removing those two tests still yields an average improvement in the objective of about 8000 for 10000 variable 10000 edge problems and QPro + Cplex finds a better solution to every problem.

The reduction ratios for QPro were very good overall and extremely good for a few problems in each size. For example, the QPro percent reduction was on average 160x greater than that achieved by Cplex on problems 4, 15, 67 and 80, and the time to best solution for QPro + Cplex was 160x faster for problems 29, 67 and 80.  Problems 67 and 80 (10000 variables) were solved to optimality by Cplex in 0.01 seconds after about 2 seconds when coupled with QPro versus 600+ seconds for the default version of Cplex.    These two problems have factors 1, 3 and 4 in common and analysis provided in the next section indicates that these three factors are the most significant for predicting percent reduction.

Table 4. Average Results for the 96 Test Runs comparing QPro+Cplex to Cplex

| | | QPro | | | Cplex | | | % | |
|---|---|---|---|---|---|---|---|---|---|
| Size | Edges | Time | Total Time | % Reduce | time | % Reduce | Time Factor | Reduce Factor | Objective Difference |
| 1000 | 5000 | 0.01 | 4 | 36 | 25 | 11 | 7 | 3 | 0 |
| 1000 | 10000 | 0.01 | 15 | 31 | 25 | 4 | 2 | 9 | 0.1 |
| 5000 | 25000 | 0.37 | 111 | 34 | 187 | 11 | 2 | 3 | 2887 |
| 5000 | 50000 | 0.30 | 157 | 29 | 136 | 3 | 1 | 11 | 465 |
| 10000 | 100000 | 1.06 | 453 | 20 | 600 | 0.4 | 1 | 56 | 159332 |
| 10000 | 500000 | 1.34 | 336 | 31 | 583 | 3 | 2 | 10 | 20712 |

Figure 5 slices the data by Problem ID (Table 2) and provides the average *Q* reduction and time factor multiple of QPro+Cplex over default Cplex. It identifies problem IDs 3, 13 and 16 as having over 50x more percent reductions and being solved 30x faster than default Cplex. These three problem types have factors 3 and 4 (high percentage of large quadratic outliers) in common. As anticipated, there is a positive correlation between percent reduction and time to best solution.

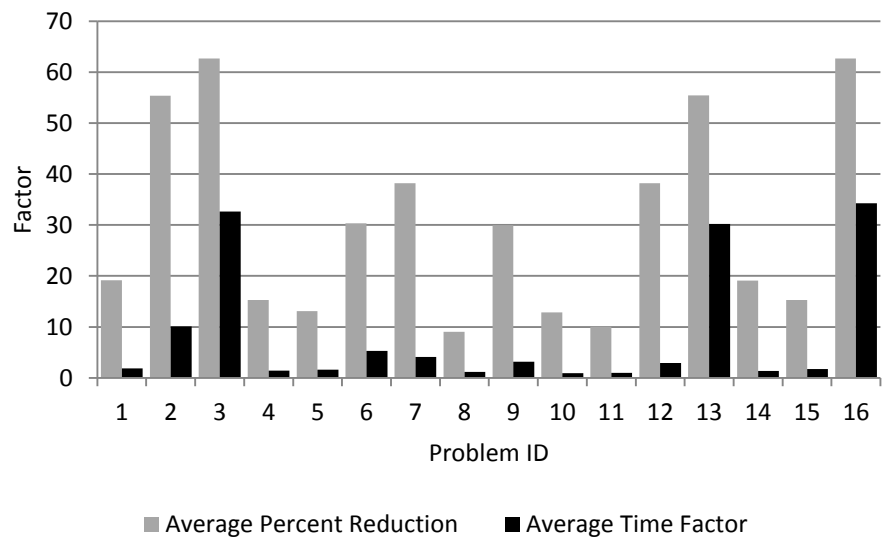

Figure 5. Average Time to Solution and Reduction Factors Categorized by Problem ID

While dramatic improvements were found when solving QUBO with non-uniform distributions, sample testing of the benchmark maxcut problems available at http://www.stanford.edu/~yyye/yyye/Gset/ could not determine value assignments for any variables using these rules because the QUBO models created for maxcut problems are very highly structured with each diagonal coefficient $c_{ii}$ equalling $-\sum\limits_{\substack{i,j \\ i \neq j}} c_{ij}$, while each nonzero $c_{ij} \in \{-1, 1\}$ and there are no positive off diagonal elements. Our rules did not yield reductions on the 10% dense and uniformly distributed ORLIB 2500 variable problems [2] due to distribution uniformity and density.

On average about four passes were made through the loop described in Section 5 Pseudocode with the number of variables fixed declining with each pass, and with the number of passes ranging from 0 to 17. For this problem set with an average reduction of 30%, Rule 3 yielded about twice the number of reductions as Rules 1 and 2 which might be expected since Rule 3 sets two variables at a time equal to one. The majority of the variables reduced were set to 1, however, the number of reductions per rule is very much dependent on the values in the $Q$ matrix.

### 6.1.2 Interpretation of Cplex Results

Results show that the primary (linear) factors most affecting percent reduction are: Magnitude of coefficient range, Size of Quadratic multiplier, and % Quadratics multiplied (factors 1, 3 and 4 in Table 1). Thus, when collecting data and modeling a problem it would be desirable to emphasize these three factors so that it is more likely that large problems can be reduced and more quickly solved.

In general, increasing the range of coefficients tended to slightly decrease the percent reduction. The explanation is that increasing the range of coefficients makes the distributions more uniform than if the range is smaller. For example, if the linear multiplier is 10x and the linear coefficient randomly generated is between [-100, 100] then there are more possibilities of not producing outliers because numbers such as 10, 20, 30, … 100 are likely not to be outliers. However, if the range is between [-10,10] then the number of outliers is increased, which allows more reductions.

Increasing the percentage of large quadratics (factors 3 and 4) tends to increase the percent reduction because it increases the use of Rule 3 that determines values for two variables at a time. It also adds $c_{ij}$ to the corresponding $c_{ii}$ and $c_{jj}$ for variables that were not determined, possibly changing them into determined variables. Factor 4 increases the percentage of large quadratics and it was the most significant factor in five of the six problem types. Table 5 summarizes the effects of the six factors on the six problem types by listing the amount change to $Q$ percent reduction when going from a low to a high factor setting. For example, Factor 1 changes the range of the uniform distribution from (-10,10) to (-100, 100) and the average effect is a slight decrease in percent reduction. Table 5 shows factor 4 (percent quadratic outliers) had the largest impact for all six problem types.

Table 5. The Primary and Interaction Effects of the 6 Factors on Percent Reduction

| Factor ID | Factor ID | Problem ID | | | | | |
|---|---|---|---|---|---|---|---|
| | | P1 | P2 | P3 | P4 | P5 | P6 |
| 1 | 1 | -4 | -3 | -2 | -2 | -3 | -2 |
| 2 | 2 | 0 | 0 | 0 | 0 | 0 | 0 |
| 3 | 3 | 2 | 8 | 3 | 7 | 6 | 20 |
| 4 | 4 | 13 | 17 | 13 | 16 | 17 | 20 |
| 5 | 5 | 0 | 0 | 0 | 0 | 0 | 0 |
| 6 | 6 | 0 | 0 | 0 | 1 | 1 | -1 |
| 1 | 2 | 0 | 0 | 0 | 0 | 0 | 0 |
| 1 | 3 | 0 | 1 | 0 | 1 | 0 | -1 |
| 1 | 4 | -1 | -2 | 0 | -1 | -1 | -1 |
| 1 | 5 | 0 | 0 | 0 | 0 | 0 | 0 |
| 1 | 6 | 1 | 3 | 1 | 2 | 2 | 20 |
| 2 | 3 | 0 | 0 | 0 | 0 | 0 | 0 |
| 2 | 4 | 0 | 0 | 0 | 0 | 0 | 0 |
| 2 | 5 | 1 | 3 | 1 | 2 | 2 | 20 |
| 2 | 6 | 0 | 0 | 0 | 0 | 0 | 0 |
| 3 | 4 | 1 | 3 | 1 | 2 | 2 | 20 |
| 3 | 5 | 0 | 0 | 0 | 0 | 0 | 0 |
| 3 | 6 | -1 | -2 | 0 | -1 | -1 | -1 |
| 4 | 5 | 0 | 0 | 0 | 0 | 0 | 0 |
| 4 | 6 | 0 | 1 | 0 | 1 | 0 | -1 |
| 5 | 6 | 0 | 0 | 0 | 0 | 0 | 0 |

Table 5 also shows there is some confounding of the interaction between factors due to the setup of the experimental design. For example, factor interactions 1-6 and 2-5 and 3-4 are confounded, meaning that they have the same test setups. An approach used to resolve confounding is to look at the primary effects of the interactions and disregard interactions having small primary effects. In this case factors 1, 2, 5 and 6 have relatively small individual effects and so we would not expect their interactions to be large. Therefore, the 20% reduction in P6 of Table 5 is most likely associated with the interaction between factors 3 and 4, both of which are individually large.

Table 5 provides data for a surface response equation generated by multiple linear regression that can be used to estimate the percent reduction that will occur when setting the six factors at a value between their defined bounds. Averaging effects and taking into account confounding and disregarding small interaction effects, an estimate of the average percent reduction for these problems is

$$\text{PR}(f_1, f_2, f_3, f_4, f_5, f_6) = -3f_1 + 8f_3 + 16f_4 + 5f_3f_4 + 30 \tag{2}$$

where the $f_i$ values are in the interval [-1, 1] and represent the range of values for the factors in Table 1.  For example, $f_1 = -1$ means the range is [-10, 10] and +1 is [-100, 100].  Intermediate values can be used to generate expected reduction amounts such as a range [-50, 50]..  The constant 30 is the average percent reduction if all factors were set to the middle of their range (implemented as $f_i = 0$).  Maximum *estimated* percent reduction occurs when the $Q$ range is [-10, 10], the quadratic multiplier is 15% and percent quadratic multiplied is 20%.

### 6.1.3  Robustness of Results  under Randomness

To support that our conclusions are based on results that were typical and not "cherry picked" or out of the ordinary, we ran repeated randomized tests on some problems.   We randomly selected test number 15 for the 1000 variable problem with 5000 and 10000 edges then generated 100 instances using a current time seeded random number generator, applied QPro and recorded the percent reduction in $Q$.   The distribution of the count of the percent reductions found is shown in Figure 6.   For the original run, the percent reductions were 18% for the 10000 edge problem and 20% for the 5000 edge problem and for the 100 random instances for test fifteen, the average number of reductions for 10000 edges was 15% ± 5% and for 5000 edges 20% ± 3%, indicating that the problems used in analysis were not out of the ordinary.  The 10000 edge problems had a wider distribution (± 5% vs ± 3%) because 10% of the problems yielded no reductions, revealing that reductions can be sensitive to random changes in $Q$.  An early article recognizing that small changes to $Q$ can have large effects on problem difficulty is that of [16].

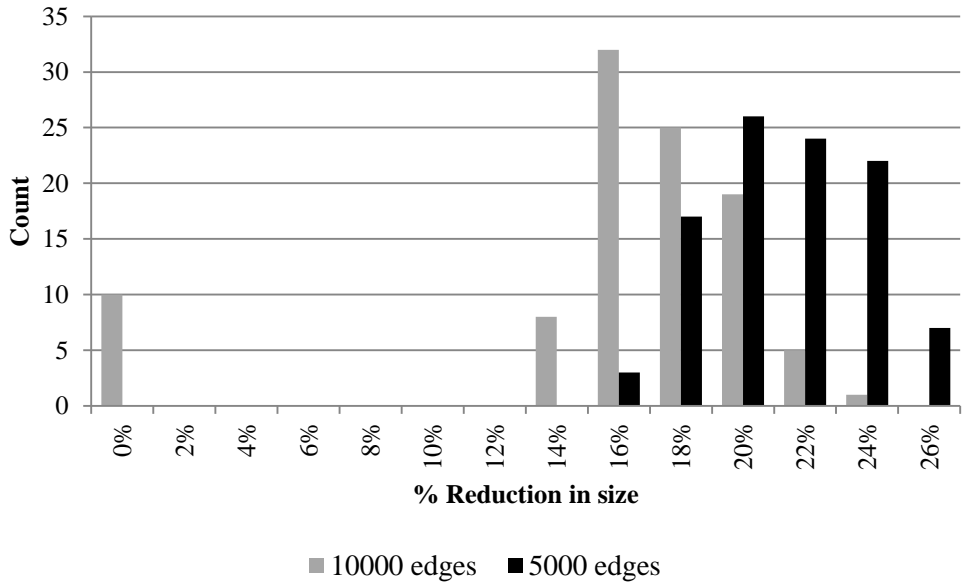

Figure 6.  Percent Reduction using Random $Q$ for Problem ID 15

Two problems (3 and 16) for the largest and densest problems (P6) showed dramatic reductions and decreases in time to best solution. After QPro these problems were solved to optimality in 0.01 seconds versus not even entering the branch-and-cut phase of Cplex after 600 seconds without QPro. For problems with these characteristic $Q$ matrices, Cplex found fewer than 40 reductions while QPro found over 7000. As a test of robustness of these results, 100 random samples of 10000 variable 500000 edge problems were generated using the characteristics for Problem IDs 3 and 16. The narrow frequency distribution of count of percent reduction shown in Figure 7 illustrates that these problem types are robust to the reduction rules and consistently yield very large reductions when random changes are made to the elements of $Q$.

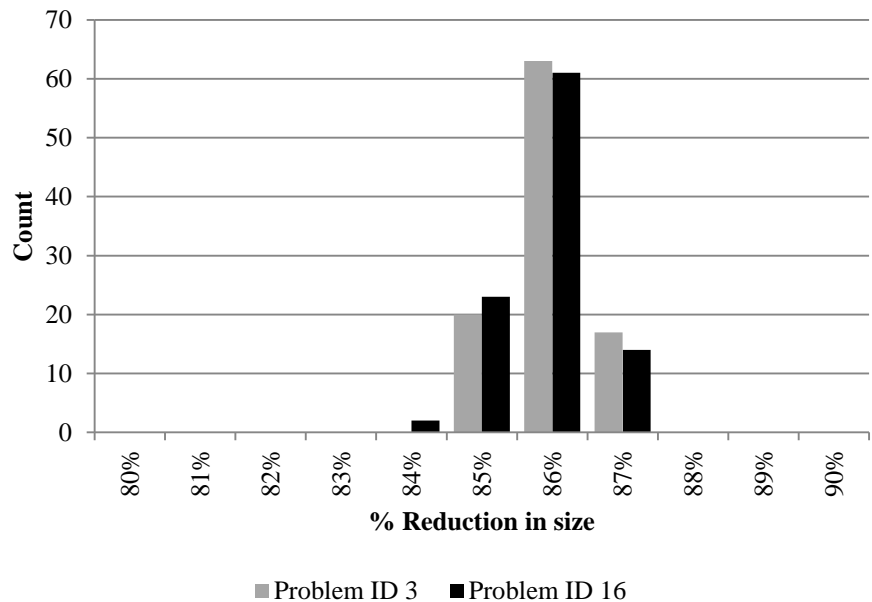

Figure 7. Percent Reduction using Random $Q$ for 10000 variables Problem IDs 3 and 16

### 6.2 Test Results Using Path Relinking Metaheuristic

The same testing approach was used with a tabu search metaheuristic with path relinking as described in [20], denoted by PR2 and made available for use in this paper. PR2 has three parameters affecting performance: run time limit, number of iterations, and tabu tenure. For each size problem, all test runs were performed in the same computing environment as with Cplex, each was given the same PR2 input parameters, and the best solution along with time to best solution recorded. The average time to best solution for each set of problems is shown in Table 6 along with the average time factor and percent objective difference. The time factor is calculated as the average of individual time factors which are the PR2 Default time divided by QPro+PR2.  The average time factor in the table is the average of the time factors for the sixteen tests for each size and density.

The averages show that QPro+PR2 was about 4x faster to a slightly better solution.   Detailed results from testing are provided in Appendix B and those results show that both approaches found the same answer for the 1000 variable problems for 31 of the 32 tests and that QPro+PR2 was over 30x faster on three of the 32 1000-variable problems.  The detailed results indicate that for the 5000 variable problems QPro+PR2 consistently had better objectives and was slightly faster.

Table 6.  Average Results for the 96 Test Runs comparing QPro+PR2 to PR2

| Size | Edges | QPro+PR2 Time | PR2 Default Time | Average Time Factor | Percent Objective Difference |
|---|---|---|---|---|---|
| 1000 | 5000 | 1.0 | 3.3 | 7 | 0 |
| 1000 | 10000 | 2.0 | 3.8 | 8 | 0 |
| 5000 | 25000 | 45.8 | 49.7 | 1 | 0.05 |
| 5000 | 50000 | 53.0 | 77.3 | 8 | 0.04 |
| 10000 | 100000 | 77.5 | 87.8 | 3 | 0.13 |
| 10000 | 500000 | 88.5 | 117.8 | 1 | 0 |
| | Averages | 45 | 57 | 4 | 0.04 |

Figure 8 averages the time factor improvements over $Q$ size and density for each of the 16 problem types, e.g., for Problem Type 2 from Table 3 (over all sizes and densities)  QPro + PR2 was about 10x faster than PR2 alone.  Figure 8 illustrates that problems of type 16 had significantly better improvements in PR2 time to solution, which is also consistent with the Cplex results.

Figure 9 drills down by problem size and shows that the majority of improvement in time was in the 1000 variable problems, which may be due to input parameter selections not being tuned for the larger problems.  The purpose of this research was not to compare heuristic and exact methods, however we found that QPro had more of an objective function value impact on Cplex than on PR2 because PR2 is already very good at quickly finding near optimal solutions.

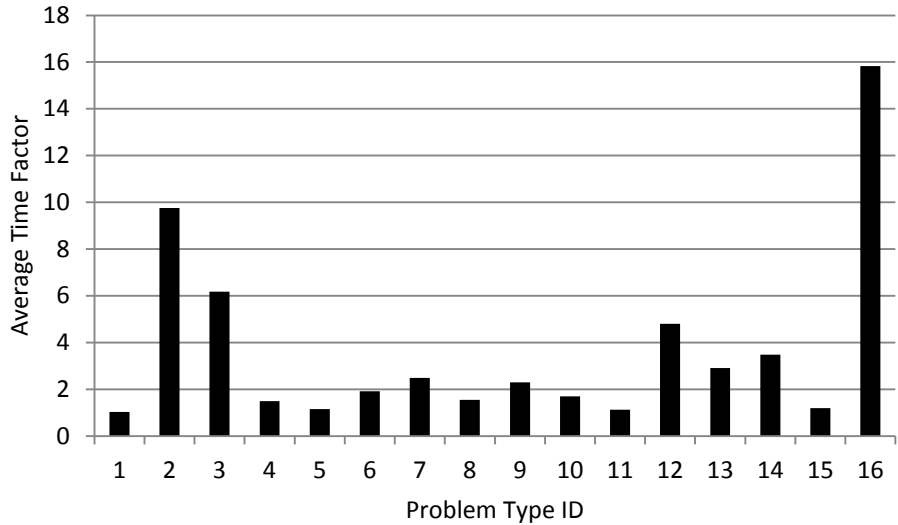

Figure 8. PR2 Time to Solution Improvements for each Problem Type

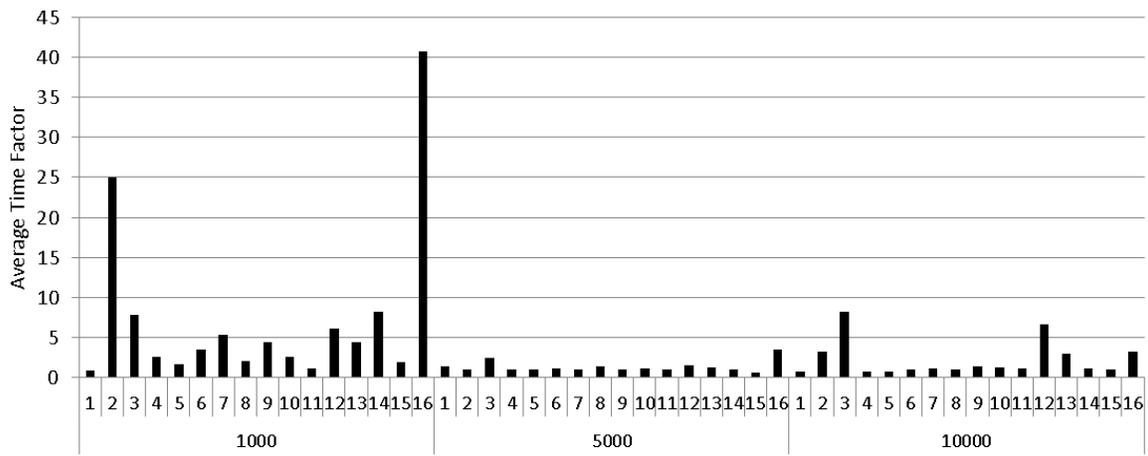

Figure 9. PR2 Time to Solution Improvements for each Problem Type by Problem Size

## 6. SUMMARY & CONCLUSIONS

In summary, this research is motivated by the fact that the QUBO framework constitutes a type of network that is useful for modeling many types of optimization problems, and by the recent development that gives QUBO a prominent role in the evaluation of modern quantum annealing computers. Accordingly, it becomes valuable to discover more effective preprocessing methods that can reduce the size of the $Q$ matrix to reduce the time to find good solutions. Our work builds on the recognition that many business problems modeled using big data are unstructured and subject to randomness, and we have accompanied our research into reducing $Q$ by developing a new set of test problems to more accurately reflect these types of models. The resulting problems have elements that are sparsely connected with the majority of $Q$ elements being uniformly distributed but with varying amounts of outlier elements.

The principal contribution of our research is the creation and justification of five rules for reducing the size of QUBO. We have presented basic pseudocode for combining the rules into a rapid preprocessor called QPro and have tested the results of using our preprocessor with the exact solver Cplex and with a tabu search metaheuristic incorporating path relinking. Careful testing and analysis shows that the $Q$ characteristics most influencing reduction are the range of the uniformly distributed elements and the number and magnitude of the quadratic outliers

In conclusion, we have established that the QPro preprocessing implementation is very fast and effective at reducing the time to obtain high quality solutions. We have additionally identified ways to apply the rules to carry out sensitivity analysis and achieve robustness, as well as identifying transformations that increase the size, but reduce edge density per node.

## ACKNOWLEDGMENTS.

The authors wish to acknowledge Professor Jeff Kennington for his inspiration and contributions to the literature of network analysis, and to thank Jin-Kao Hao and Wang Yang for providing the PR2 QUBO metaheuristic which has logged the fastest times to obtain high quality QUBO solutions. Finally, we express our gratitude to Steve Reinhardt and Michael Booth of D-Wave Systems for acquainting us with the issues of minor embedding of graphs onto their quantum annealing Chimera graph structure.

# REFERENCES

[1] W. P. Adams and P. M. Dearing, "On the equivalence between roof duality and Lagrangian duality for unconstrained 0–1 quadratic programming problems," *Discrete Applied Mathematics*, vol. 48, no. 1, pp. 1-20, 1994.

[2] J. E. Beasley. OR-Library. [Online]. http://people.brunel.ac.uk/~mastjjb/jeb/orlib/bqpinfo.html

[3] Dimitris Bertsimas, Omid Nohadani, and K. M. Teo, "Robust optimization for unconstrained simulation-based problems," *Operations Research*, vol. 68, no. 1, pp. 161-178, 2010.

[4] Sergio Boixo et al., "Evidence for quantum annealing with more than one hundred qubits," *Nature Physics*, vol. 10, pp. 218-224, 2014.

[5] Tomas Boothby, Andrew D. King, and Aiden Roy, "Fast clique minor generation in Chimera qubit connectivity graphs," *Quantum Information Processing*, vol. 15, no. 1, pp. 495-508, 2016.

[6] Endre Boros, Peter L. Hammer, and Gabriel Tavares, "Preprocessing of unconstrained quadratic binary optimization," *Rutcor Research Report RRR 10–2006*, pp. 1-54, 2006.

[7] Vicky Choi, "Minor-embedding in adiabatic quantum computation: I. The parameter setting problem," *Quantum Information Processing*, vol. 7, pp. 193-209, 2008.

[8] Fred Glover, Gary Kochenberger, and Bahram Alidaee, "Adaptive memory tabu search for binary quadratic programs," *Management Science*, vol. 44, no. 3, pp. 336-345, 1998.

[9] Jeffery Kennington and Karen Lewis, "Generalized networks: The theory of preprocessing and an empirical analysis," *INFORMS Journal on Computing*, vol. 16, no. 2, pp. 0162-0173, 2004.

[10] Jongkwang Kim and Thomas Wilhelm, "What is a complex graph?," *Physica A: Statistical Mechanics and its Applications*, vol. 387, no. 11, pp. 2637-2652, 2008.

[11] Gary Kochenberger, Fred Glover, Bahram Alidaee, and Cesar Rego, "A unified modeling and solution framework for combinatorial optimization problems," *OR Spectrum*, vol. 26, no. 3, pp. 237-250, 2004.

[12] Gary Kochenberger et al., "The unconstrained binary quadratic: a survey," *Journal of Combinatorial Optimization*, vol. 28, no. 1, pp. 55-81, 2014.

[13] D Laughhunn, "Quadratic binary programming with application to capital-budgeting problems," *Operations Research*, vol. 18, no. 3, pp. 454-461, 1970.

[14] Andrew Lucas, "Ising formulations of many NP Problems," *Frontiers in Physics*, vol. 5, no. arXiv:1302.5843, p. 2, 2014.

[15] D. C. Montgomery, *Design and Analysis of Experiments*. New York: John Wiley and Sons, 1984.

[16] P. M. Pardalos and G. P. Rodgers, "Computational aspects of a branch and bound algorithm for quadratic zero-one programming," *Computing*, vol. 45, no. 2, pp. 131-144, 1990.

[17] Gili Rosenberg, Mohammad Vazifeh, Brad Woods, and Eldad Haber, "Building an iterative heuristic solver for a quantum annealer," *Computational Optimization and Applications*, pp. 1-25, 2016.

[18] G. Taguchi, A. Elsayed, and T. Hsiang, *Quality Engineering in Production Systems*. New York: McGraw-Hill, 1989.

[19] D. Wang and R. Kleinberg, "Analyzing quadratic unconstrained binary optimization problems via multicommodity flows," *Discrete Applied Mathematics*, vol. 157, no. 18, pp. 3746-3753, 2009.

[20] Yang Wang, Zhipeng Lu, Fred Glover, and Jin-Kao Hao, "Path relinking for unconstrained binary quadratic programming," *European Journal of Operational Research*, vol. 223, no. 3, pp. 595-604, 2012.

APPENDICES

A. Problem Characteristics and Detailed Test Results Using QPro and Cplex

| ID | DOE ID | Size | Density % | QPro Obj | Cplex Obj | Difference |
|---|---|---|---|---|---|---|
| 1 | 1 | 1000 | 2 | 43804 | 43804 | 0 |
| 2 | 2 | 1000 | 2 | 903359 | 903359 | 0 |
| 3 | 3 | 1000 | 2 | 96986 | 96986 | 0 |
| 4 | 4 | 1000 | 2 | 382632 | 382632 | 0 |
| 5 | 5 | 1000 | 2 | 29905 | 29905 | 0 |
| 6 | 6 | 1000 | 2 | 479397 | 479397 | 0 |
| 7 | 7 | 1000 | 2 | 55874 | 55874 | 0 |
| 8 | 8 | 1000 | 2 | 254819 | 254819 | 0 |
| 9 | 9 | 1000 | 2 | 481453 | 481453 | 0 |
| 10 | 10 | 1000 | 2 | 29668 | 29668 | 0 |
| 11 | 11 | 1000 | 2 | 260325 | 260325 | 0 |
| 12 | 12 | 1000 | 2 | 54874 | 54874 | 0 |
| 13 | 13 | 1000 | 2 | 879654 | 879652 | 2 |
| 14 | 14 | 1000 | 2 | 43543 | 43543 | 0 |
| 15 | 15 | 1000 | 2 | 382945 | 382945 | 0 |
| 16 | 16 | 1000 | 2 | 97345 | 97345 | 0 |
| 17 | 1 | 1000 | 1 | 26074 | 26074 | 0 |
| 18 | 2 | 1000 | 1 | 488919 | 488919 | 0 |
| 19 | 3 | 1000 | 1 | 52493 | 52493 | 0 |
| 20 | 4 | 1000 | 1 | 215402 | 215402 | 0 |
| 21 | 5 | 1000 | 1 | 18662 | 18662 | 0 |
| 22 | 6 | 1000 | 1 | 277851 | 277851 | 0 |
| 23 | 7 | 1000 | 1 | 32259 | 32259 | 0 |
| 24 | 8 | 1000 | 1 | 156133 | 156133 | 0 |
| 25 | 9 | 1000 | 1 | 273774 | 273774 | 0 |
| 26 | 10 | 1000 | 1 | 18405 | 18405 | 0 |
| 27 | 11 | 1000 | 1 | 158849 | 158849 | 0 |
| 28 | 12 | 1000 | 1 | 32406 | 32406 | 0 |
| 29 | 13 | 1000 | 1 | 484897 | 484897 | 0 |
| 30 | 14 | 1000 | 1 | 25883 | 25883 | 0 |
| 31 | 15 | 1000 | 1 | 216039 | 216039 | 0 |
| 32 | 16 | 1000 | 1 | 52698 | 52698 | 0 |

A1. 1000 variable problems size, density and objective values

| ID | Time | QPro | | Cplex | | Ratios | |
|---|---|---|---|---|---|---|---|
| | | Total Time | % Reduce | Time | % Reduce | Time | Reduce |
| 1 | 0.01 | 8 | 20.4 | 20 | 2.3 | 2.5 | 9 |
| 2 | 0.01 | 3 | 54.8 | 10 | 11.7 | 3.3 | 5 |
| 3 | 0.02 | 2 | 62.8 | 9 | 3.6 | 4.5 | 17 |
| 4 | 0.01 | 20 | 18.8 | 65 | 0.2 | 3.3 | 94 |
| 5 | 0.01 | 10 | 10.9 | 20 | 0.6 | 2.0 | 18 |
| 6 | 0.02 | 7 | 32.7 | 15 | 1.2 | 2.1 | 27 |
| 7 | 0.02 | 5 | 43.3 | 11 | 8.1 | 2.2 | 5 |
| 8 | 0.01 | 45 | 5.6 | 50 | 1.5 | 1.1 | 4 |
| 9 | 0.02 | 7 | 32.1 | 15 | 2.1 | 2.1 | 15 |
| 10 | 0.01 | 14 | 10.8 | 18 | 0.3 | 1.3 | 36 |
| 11 | 0.01 | 80 | 6.9 | 45 | 3.1 | 0.6 | 2 |
| 12 | 0.01 | 3 | 42.6 | 11 | 7.2 | 3.7 | 6 |
| 13 | 0.03 | 3 | 55.6 | 7 | 10.0 | 2.3 | 6 |
| 14 | 0.01 | 9 | 20.4 | 22 | 2.2 | 2.4 | 9 |
| 15 | 0.01 | 20 | 18.8 | 70 | 0.2 | 3.5 | 94 |
| 16 | 0.02 | 2 | 62.9 | 11 | 3.9 | 5.5 | 16 |
| 17 | 0.01 | 5 | 27.9 | 8 | 10.8 | 1.6 | 3 |
| 18 | 0.01 | 3 | 47.2 | 6 | 11.7 | 2.0 | 4 |
| 19 | 0.01 | 2 | 56.5 | 3 | 16.0 | 1.5 | 4 |
| 20 | 0.01 | 7 | 19.9 | 14 | 4.4 | 2.0 | 5 |
| 21 | 0.01 | 5 | 24.0 | 8 | 7.5 | 1.6 | 3 |
| 22 | 0.01 | 2 | 41.3 | 12 | 5.4 | 6.0 | 8 |
| 23 | 0.02 | 0.2 | 51.7 | 1 | 24.4 | 5.0 | 2 |
| 24 | 0.01 | 7 | 16.7 | 12 | 5.7 | 1.7 | 3 |
| 25 | 0.01 | 3 | 40.1 | 11 | 5.6 | 3.7 | 7 |
| 26 | 0.01 | 3 | 24.1 | 2 | 7.0 | 0.7 | 3 |
| 27 | 0.01 | 7 | 18.1 | 12 | 6.7 | 1.7 | 3 |
| 28 | 0.02 | 0.1 | 51.8 | 0.5 | 23.3 | 5.0 | 2 |
| 29 | 0.01 | 2 | 47.0 | 280 | 10.3 | 140.0 | 5 |
| 30 | 0.01 | 5 | 28.1 | 8 | 12.1 | 1.6 | 2 |
| 31 | 0.01 | 6 | 19.8 | 12 | 4.4 | 2.0 | 5 |
| 32 | 0.01 | 2 | 56.7 | 4 | 16.7 | 2.0 | 3 |
| Averages | 0.01 | 9 | 33 | 25 | 7 | 7 | 13 |

A2.  1000 variable time to best solution and percent reductions

| ID | DOE ID | Size | Density % | QPro Obj | Cplex Obj | Difference |
|---|---|---|---|---|---|---|
| 33 | 1 | 5000 | 0.4 | 204391 | 204262 | 129 |
| 34 | 2 | 5000 | 0.4 | 4230341 | 4227909 | 2432 |
| 35 | 3 | 5000 | 0.4 | 448027 | 448000 | 27 |
| 36 | 4 | 5000 | 0.4 | 1793275 | 1767620 | 25655 |
| 37 | 5 | 5000 | 0.4 | 142924 | 142839 | 85 |
| 38 | 6 | 5000 | 0.4 | 2277417 | 2276055 | 1362 |
| 39 | 7 | 5000 | 0.4 | 256085 | 256066 | 19 |
| 40 | 8 | 5000 | 0.4 | 1220530 | 1219811 | 719 |
| 41 | 9 | 5000 | 0.4 | 2282365 | 2278151 | 4214 |
| 42 | 10 | 5000 | 0.4 | 142568 | 142465 | 103 |
| 43 | 11 | 5000 | 0.4 | 1247769 | 1247377 | 392 |
| 44 | 12 | 5000 | 0.4 | 259523 | 259489 | 34 |
| 45 | 13 | 5000 | 0.4 | 4175231 | 4172468 | 2763 |
| 46 | 14 | 5000 | 0.4 | 201192 | 201035 | 157 |
| 47 | 15 | 5000 | 0.4 | 1798995 | 1790940 | 8055 |
| 48 | 16 | 5000 | 0.4 | 450303 | 450260 | 43 |
| 49 | 1 | 5000 | 0.2 | 118508 | 118430 | 78 |
| 50 | 2 | 5000 | 0.2 | 2482738 | 2482738 | 0 |
| 51 | 3 | 5000 | 0.2 | 246707 | 246707 | 0 |
| 52 | 4 | 5000 | 0.2 | 1091258 | 1088083 | 3175 |
| 53 | 5 | 5000 | 0.2 | 88070 | 87970 | 100 |
| 54 | 6 | 5000 | 0.2 | 1360353 | 1360299 | 54 |
| 55 | 7 | 5000 | 0.2 | 149308 | 149308 | 0 |
| 56 | 8 | 5000 | 0.2 | 789052 | 788540 | 512 |
| 57 | 9 | 5000 | 0.2 | 1357177 | 1357177 | 0 |
| 58 | 10 | 5000 | 0.2 | 87401 | 87308 | 93 |
| 59 | 11 | 5000 | 0.2 | 830412 | 829672 | 740 |
| 60 | 12 | 5000 | 0.2 | 151219 | 151219 | 0 |
| 61 | 13 | 5000 | 0.2 | 2390595 | 2390595 | 0 |
| 62 | 14 | 5000 | 0.2 | 117471 | 117349 | 122 |
| 63 | 15 | 5000 | 0.2 | 1090106 | 1087537 | 2569 |
| 64 | 16 | 5000 | 0.2 | 248397 | 248397 | 0 |

A3.  5000 variable problems size, density and objective values

| | QPro | | | Cplex | | Ratios | |
|---|---|---|---|---|---|---|---|
| ID | Time | Total Time | % Reduce | Time | % Reduce | Time | Reduce |
| 33 | 0.24 | 270 | 18.8 | 90 | 2.2 | 0.3 | 8 |
| 34 | 0.38 | 100 | 52.6 | 275 | 8.2 | 2.7 | 7 |
| 35 | 0.4 | 22 | 56.5 | 180 | 3.4 | 8.0 | 19 |
| 36 | 0.19 | 280 | 16.1 | 60 | 0.4 | 0.2 | 43 |
| 37 | 0.2 | 45 | 9.5 | 70 | 0.4 | 1.5 | 17 |
| 38 | 0.38 | 290 | 31.6 | 120 | 1.0 | 0.4 | 31 |
| 39 | 0.38 | 30 | 40.2 | 200 | 5.5 | 6.6 | 8 |
| 40 | 0.16 | 45 | 7.0 | 60 | 1.1 | 1.3 | 6 |
| 41 | 0.38 | 280 | 31.4 | 50 | 0.9 | 0.2 | 32 |
| 42 | 0.2 | 290 | 9.5 | 50 | 0.4 | 0.2 | 20 |
| 43 | 0.19 | 50 | 8.0 | 60 | 1.9 | 1.2 | 4 |
| 44 | 0.41 | 116 | 40.9 | 190 | 5.7 | 1.6 | 8 |
| 45 | 0.38 | 110 | 52.1 | 295 | 6.8 | 2.7 | 9 |
| 46 | 0.23 | 273 | 18.6 | 85 | 2.2 | 0.3 | 9 |
| 47 | 0.19 | 290 | 16.0 | 200 | 0.4 | 0.7 | 45 |
| 48 | 0.42 | 20 | 56.8 | 190 | 3.4 | 9.3 | 18 |
| 49 | 0.3 | 72 | 26.3 | 275 | 11.6 | 3.8 | 2 |
| 50 | 0.5 | 10 | 49.4 | 240 | 14.5 | 25.3 | 4 |
| 51 | 0.49 | 20 | 54.1 | 130 | 15.5 | 6.3 | 3 |
| 52 | 0.25 | 300 | 21.5 | 30 | 5.8 | 0.1 | 3 |
| 53 | 0.39 | 80 | 22.1 | 100 | 9.8 | 1.2 | 2 |
| 54 | 0.4 | 15 | 42.3 | 295 | 7.5 | 19.2 | 5 |
| 55 | 0.45 | 20 | 48.4 | 110 | 18.4 | 5.4 | 3 |
| 56 | 0.23 | 280 | 17.5 | 210 | 6.9 | 0.7 | 2 |
| 57 | 0.4 | 45 | 42.1 | 299 | 7.6 | 6.6 | 5 |
| 58 | 0.38 | 95 | 21.3 | 100 | 9.1 | 1.0 | 2 |
| 59 | 0.27 | 290 | 18.4 | 180 | 7.8 | 0.6 | 2 |
| 60 | 0.45 | 20 | 48.0 | 122 | 18.3 | 6.0 | 3 |
| 61 | 0.41 | 32 | 48.8 | 295 | 13.0 | 9.1 | 4 |
| 62 | 0.31 | 107 | 26.1 | 140 | 11.6 | 1.3 | 2 |
| 63 | 0.25 | 280 | 21.6 | 280 | 5.7 | 1.0 | 3 |
| 64 | 0.49 | 18 | 54.5 | 130 | 16.3 | 7.0 | 4 |
| Averages | 0.3 | 131 | 32 | 160 | 7 | 4 | 10 |

A4.  5000 variable time to best solution and percent reductions

| ID | DOE ID | Size | Density % | QPro Obj | Cplex Obj | Difference |
|---|---|---|---|---|---|---|
| 65 | 1 | 10000 | 1 | 1555611 | 1555611 | 0 |
| 66 | 2 | 10000 | 1 | 37838000 | 37833600 | 4400 |
| 67 | 3 | 10000 | 1 | 4212259 | 4212053 | 206 |
| 68 | 4 | 10000 | 1 | 10947337 | 10947300 | 37 |
| 69 | 5 | 10000 | 1 | 985534 | 985534 | 0 |
| 70 | 6 | 10000 | 1 | 17759300 | 17759300 | 0 |
| 71 | 7 | 10000 | 1 | 2249326 | 2249326 | 0 |
| 72 | 8 | 10000 | 1 | 5495629 | 5495629 | 0 |
| 73 | 9 | 10000 | 1 | 17743900 | 17743900 | 0 |
| 74 | 10 | 10000 | 1 | 981258 | 981258 | 0 |
| 75 | 11 | 10000 | 1 | 5596983 | 5596983 | 0 |
| 76 | 12 | 10000 | 1 | 2263103 | 2263103 | 0 |
| 77 | 13 | 10000 | 1 | 37181848 | 36855100 | 326748 |
| 78 | 14 | 10000 | 1 | 1547867 | 1547867 | 0 |
| 79 | 15 | 10000 | 1 | 10968200 | 10968200 | 0 |
| 80 | 16 | 10000 | 1 | 4224394 | 4224394 | 0 |
| 81 | 1 | 10000 | 0.2 | 429901 | 426791 | 3110 |
| 82 | 2 | 10000 | 0.2 | 9028291 | 9020055 | 8236 |
| 83 | 3 | 10000 | 0.2 | 954647 | 954517 | 130 |
| 84 | 4 | 10000 | 0.2 | 3533451 | 2317043 | 1216408 |
| 85 | 5 | 10000 | 0.2 | 299673 | 299589 | 84 |
| 86 | 6 | 10000 | 0.2 | 4651197 | 4646122 | 5075 |
| 87 | 7 | 10000 | 0.2 | 542974 | 542933 | 41 |
| 88 | 8 | 10000 | 0.2 | 2429373 | 2387583 | 41790 |
| 89 | 9 | 10000 | 0.2 | 4661455 | 4656335 | 5120 |
| 90 | 10 | 10000 | 0.2 | 296544 | 296456 | 88 |
| 91 | 11 | 10000 | 0.2 | 2544042 | 2506695 | 37347 |
| 92 | 12 | 10000 | 0.2 | 547371 | 547344 | 27 |
| 93 | 13 | 10000 | 0.2 | 8696580 | 8685361 | 11219 |
| 94 | 14 | 10000 | 0.2 | 426669 | 423492 | 3177 |
| 95 | 15 | 10000 | 0.2 | 3529275 | 2311948 | 1217327 |
| 96 | 16 | 10000 | 0.2 | 961129 | 960992 | 137 |

A5.  10000 variable problems size, density and objective values

| ID | QPro | | | Cplex | | Ratios | |
|---|---|---|---|---|---|---|---|
| | Time | Total Time | % Reduce | time | % Reduce | Time | Reduce |
| 65 | 0.85 | 601 | 1.4 | 600 | 0.2 | 1.0 | 6 |
| 66 | 3.13 | 23 | 74.3 | 600 | 1.2 | 25.9 | 63 |
| 67 | 3.53 | 4 | 85.5 | 600 | 0.4 | 169.5 | 237 |
| 68 | 0.27 | 600 | 0.1 | 600 | 0.0 | 1.0 | 3 |
| 69 | 0.14 | 600 | 0.2 | 600 | 0.1 | 1.0 | 3 |
| 70 | 0.14 | 600 | 0.1 | 600 | 0.1 | 1.0 | 1 |
| 71 | 0.29 | 600 | 1.1 | 600 | 0.8 | 1.0 | 1 |
| 72 | 0.13 | 600 | 0.1 | 600 | 0.2 | 1.0 | 0 |
| 73 | 0.14 | 600 | 0.1 | 600 | 0.2 | 1.0 | 0 |
| 74 | 0.14 | 600 | 0.0 | 600 | 0.1 | 1.0 | 1 |
| 75 | 0.14 | 600 | 0.6 | 600 | 0.3 | 1.0 | 2 |
| 76 | 0.42 | 600 | 0.9 | 600 | 0.7 | 1.0 | 1 |
| 77 | 3.3 | 22 | 75.5 | 600 | 1.0 | 26.9 | 76 |
| 78 | 0.7 | 601 | 1.7 | 600 | 0.2 | 1.0 | 8 |
| 79 | 0.27 | 600 | 0.1 | 600 | 0.0 | 1.0 | 6 |
| 80 | 3.38 | 3 | 84.6 | 600 | 0.4 | 177.0 | 217 |
| 81 | 1.12 | 301 | 20.2 | 600 | 2.8 | 2.0 | 7 |
| 82 | 1.76 | 502 | 53.9 | 600 | 10.6 | 1.2 | 5 |
| 83 | 2.04 | 102 | 60.5 | 600 | 2.8 | 5.9 | 22 |
| 84 | 0.76 | 301 | 15.5 | 600 | 0.5 | 2.0 | 32 |
| 85 | 0.97 | 271 | 11.9 | 600 | 0.7 | 2.2 | 17 |
| 86 | 1.72 | 194 | 34.2 | 600 | 0.7 | 3.1 | 52 |
| 87 | 1.76 | 102 | 44.7 | 500 | 6.7 | 4.9 | 7 |
| 88 | 0.64 | 601 | 7.4 | 600 | 1.0 | 1.0 | 8 |
| 89 | 1.72 | 112 | 34.2 | 600 | 0.7 | 5.4 | 46 |
| 90 | 0.84 | 401 | 11.5 | 600 | 0.4 | 1.5 | 27 |
| 91 | 0.78 | 601 | 8.3 | 600 | 2.3 | 1.0 | 4 |
| 92 | 1.89 | 552 | 45.1 | 600 | 5.6 | 1.1 | 8 |
| 93 | 1.6 | 602 | 53.6 | 600 | 8.6 | 1.0 | 6 |
| 94 | 1.1 | 371 | 19.8 | 600 | 3.0 | 1.6 | 7 |
| 95 | 0.77 | 281 | 15.4 | 600 | 0.3 | 2.1 | 48 |
| 96 | 2.04 | 87 | 60.4 | 430 | 3.4 | 4.9 | 18 |
| | 1 | 395 | 26 | 592 | 2 | 14 | 29 |

A6. 10000 variable time to best solution and percent reductions

APPENDIX B.  Detailed Test Results using QPro and  Path Relinking Metaheuristic PR2

| ID | Size | Density % | QPro + PR2 | | PR 2 default | | Objective Difference | Time Factor |
|---|---|---|---|---|---|---|---|---|
| | | | Objective | Time | Objective | Time | | |
| 1 | 1000 | 2 | 43804 | 0.9 | 43804 | 0.9 | 0 | 1.0 |
| 2 | 1000 | 2 | 903359 | 0.3 | 903359 | 12.6 | 0 | 40.6 |
| 3 | 1000 | 2 | 96986 | 0.0 | 96986 | 0.3 | 0 | 7.5 |
| 4 | 1000 | 2 | 382632 | 6.0 | 382632 | 2.4 | 0 | 0.4 |
| 5 | 1000 | 2 | 29905 | 6.0 | 29905 | 4.7 | 0 | 0.8 |
| 6 | 1000 | 2 | 479397 | 2.0 | 479697 | 2.8 | -300 | 1.4 |
| 7 | 1000 | 2 | 55874 | 0.5 | 55874 | 4.7 | 0 | 9.0 |
| 8 | 1000 | 2 | 254819 | 2.0 | 254819 | 3.9 | 0 | 1.9 |
| 9 | 1000 | 2 | 481453 | 1.0 | 481453 | 3.9 | 0 | 3.8 |
| 10 | 1000 | 2 | 29668 | 2.0 | 29668 | 7.7 | 0 | 3.8 |
| 11 | 1000 | 2 | 260325 | 3.7 | 260325 | 5.9 | 0 | 1.6 |
| 12 | 1000 | 2 | 54874 | 0.1 | 54874 | 1.0 | 0 | 9.1 |
| 13 | 1000 | 2 | 879654 | 0.3 | 879654 | 1.7 | 0 | 6.1 |
| 14 | 1000 | 2 | 43543 | 0.6 | 43543 | 1.1 | 0 | 1.8 |
| 15 | 1000 | 2 | 382945 | 2.3 | 382945 | 4.9 | 0 | 2.1 |
| 16 | 1000 | 2 | 97345 | 0.1 | 97345 | 2.4 | 0 | 34.3 |
| 17 | 1000 | 1 | 26074 | 1.3 | 26074 | 1.0 | 0 | 0.8 |
| 18 | 1000 | 1 | 488919 | 0.5 | 488919 | 4.7 | 0 | 9.2 |
| 19 | 1000 | 1 | 52493 | 0.1 | 52493 | 0.9 | 0 | 8.2 |
| 20 | 1000 | 1 | 215402 | 1.0 | 215402 | 4.8 | 0 | 4.8 |
| 21 | 1000 | 1 | 18662 | 1.0 | 18662 | 2.6 | 0 | 2.6 |
| 22 | 1000 | 1 | 277851 | 1.9 | 277851 | 10.8 | 0 | 5.7 |
| 23 | 1000 | 1 | 32259 | 0.3 | 32259 | 0.5 | 0 | 1.6 |
| 24 | 1000 | 1 | 156133 | 1.4 | 156133 | 3.3 | 0 | 2.3 |
| 25 | 1000 | 1 | 273774 | 1.6 | 273774 | 8.1 | 0 | 5.0 |
| 26 | 1000 | 1 | 18405 | 1.7 | 18405 | 2.3 | 0 | 1.3 |
| 27 | 1000 | 1 | 158849 | 3.0 | 158849 | 2.5 | 0 | 0.8 |
| 28 | 1000 | 1 | 32406 | 0.3 | 32406 | 1.0 | 0 | 3.1 |
| 29 | 1000 | 1 | 484897 | 1.0 | 484897 | 2.7 | 0 | 2.7 |
| 30 | 1000 | 1 | 25883 | 0.2 | 25883 | 2.5 | 0 | 14.7 |
| 31 | 1000 | 1 | 216039 | 1.0 | 216039 | 1.8 | 0 | 1.8 |
| 32 | 1000 | 1 | 52698 | 0.1 | 52698 | 3.3 | 0 | 47.1 |

B1.  PR2 comparison of Objective Value and Time using the 96 Test Problems

| ID | Size | Density % | QPro + PR2 | | PR 2 default | | Objective Difference | Time Factor |
|---|---|---|---|---|---|---|---|---|
| | | | Objective | Time | Objective | Time | | |
| 33 | 5000 | 0.4 | 204411 | 43.2 | 204374 | 54.0 | 37 | 1.2 |
| 34 | 5000 | 0.4 | 4230341 | 62.4 | 4228816 | 46.0 | 1525 | 0.7 |
| 35 | 5000 | 0.4 | 448027 | 21.4 | 447965 | 58.0 | 62 | 2.7 |
| 36 | 5000 | 0.4 | 1794446 | 55.2 | 1792891 | 53.0 | 1555 | 1.0 |
| 37 | 5000 | 0.4 | 143042 | 69.2 | 142940 | 56.0 | 102 | 0.8 |
| 38 | 5000 | 0.4 | 2278847 | 56.4 | 2278361 | 53.0 | 486 | 0.9 |
| 39 | 5000 | 0.4 | 256083 | 61.4 | 256027 | 42.0 | 56 | 0.7 |
| 40 | 5000 | 0.4 | 1223703 | 74.2 | 1221522 | 50.0 | 2181 | 0.7 |
| 41 | 5000 | 0.4 | 2283311 | 61.4 | 2282826 | 40.0 | 485 | 0.7 |
| 42 | 5000 | 0.4 | 142645 | 63.2 | 142607 | 56.0 | 38 | 0.9 |
| 43 | 5000 | 0.4 | 1251370 | 63.2 | 1250671 | 50.0 | 699 | 0.8 |
| 44 | 5000 | 0.4 | 259510 | 31.4 | 259458 | 44.0 | 52 | 1.4 |
| 45 | 5000 | 0.4 | 4175186 | 47.4 | 4174697 | 45.0 | 489 | 0.9 |
| 46 | 5000 | 0.4 | 201237 | 66.2 | 201205 | 53.0 | 32 | 0.8 |
| 47 | 5000 | 0.4 | 1801032 | 67.2 | 1799503 | 54.0 | 1529 | 0.8 |
| 48 | 5000 | 0.4 | 450303 | 8.4 | 450267 | 41.0 | 36 | 4.9 |
| 49 | 5000 | 0.2 | 118480 | 52.3 | 118425 | 86.0 | 55 | 1.6 |
| 50 | 5000 | 0.2 | 2482694 | 65.5 | 2481641 | 88.0 | 1053 | 1.3 |
| 51 | 5000 | 0.2 | 246705 | 35.5 | 246695 | 80.0 | 10 | 2.3 |
| 52 | 5000 | 0.2 | 1089450 | 66.3 | 1088571 | 79.0 | 879 | 1.2 |
| 53 | 5000 | 0.2 | 88046 | 63.4 | 88024 | 81.0 | 22 | 1.3 |
| 54 | 5000 | 0.2 | 1359965 | 59.4 | 1358681 | 81.0 | 1284 | 1.4 |
| 55 | 5000 | 0.2 | 149300 | 58.5 | 149285 | 81.0 | 15 | 1.4 |
| 56 | 5000 | 0.2 | 789690 | 41.2 | 788841 | 88.0 | 849 | 2.1 |
| 57 | 5000 | 0.2 | 1356638 | 65.4 | 1356338 | 86.0 | 300 | 1.3 |
| 58 | 5000 | 0.2 | 87361 | 57.4 | 87372 | 85.0 | -11 | 1.5 |
| 59 | 5000 | 0.2 | 830362 | 63.3 | 829205 | 79.0 | 1157 | 1.2 |
| 60 | 5000 | 0.2 | 151215 | 44.5 | 151187 | 79.0 | 28 | 1.8 |
| 61 | 5000 | 0.2 | 2390567 | 45.4 | 2389123 | 74.0 | 1444 | 1.6 |
| 62 | 5000 | 0.2 | 117435 | 63.3 | 117438 | 77.0 | -3 | 1.2 |
| 63 | 5000 | 0.2 | 1088555 | 63.3 | 1086962 | 25.0 | 1593 | 0.4 |
| 64 | 5000 | 0.2 | 248397 | 33.5 | 248375 | 71.0 | 22 | 2.1 |

B1 (continued).  PR2 comparison of Objective Value and Time using the 96 Test Problems

| ID | Size | Density % | QPro + PR2 | | PR 2 default | | Objective Difference | Time Factor |
|---|---|---|---|---|---|---|---|---|
| | | | Objective | Time | Objective | Time | | |
| 65 | 10000 | 1 | 1620836 | 118.9 | 1620899 | 83.0 | -63 | 0.7 |
| 66 | 10000 | 1 | 37837990 | 7.1 | 37837887 | 40.0 | 103 | 5.6 |
| 67 | 10000 | 1 | 4212259 | 3.9 | 4212259 | 58.0 | 0 | 14.8 |
| 68 | 10000 | 1 | 12474795 | 113.3 | 12472295 | 90.0 | 2500 | 0.8 |
| 69 | 10000 | 1 | 1054227 | 139.1 | 1054083 | 99.0 | 144 | 0.7 |
| 70 | 10000 | 1 | 18214481 | 120.1 | 18214499 | 123.0 | -18 | 1.0 |
| 71 | 10000 | 1 | 2267676 | 31.3 | 2267676 | 29.0 | 0 | 0.9 |
| 72 | 10000 | 1 | 7276904 | 92.1 | 7283019 | 101.0 | -6115 | 1.1 |
| 73 | 10000 | 1 | 18200291 | 62.1 | 18200834 | 132.0 | -543 | 2.1 |
| 74 | 10000 | 1 | 1048779 | 71.1 | 1048929 | 130.0 | -150 | 1.8 |
| 75 | 10000 | 1 | 7469937 | 114.1 | 7471912 | 131.0 | -1975 | 1.1 |
| 76 | 10000 | 1 | 2278831 | 2.4 | 2278831 | 30.0 | 0 | 12.4 |
| 77 | 10000 | 1 | 37181848 | 25.3 | 37181832 | 141.0 | 16 | 5.6 |
| 78 | 10000 | 1 | 1612715 | 74.7 | 1612817 | 103.0 | -102 | 1.4 |
| 79 | 10000 | 1 | 12487346 | 80.3 | 12480960 | 99.0 | 6386 | 1.2 |
| 80 | 10000 | 1 | 4224394 | 3.4 | 4224394 | 16.0 | 0 | 4.7 |
| 81 | 10000 | 0.2 | 430087 | 151.1 | 430031 | 131.0 | 56 | 0.9 |
| 82 | 10000 | 0.2 | 9029102 | 105.8 | 9006658 | 108.0 | 22444 | 1.0 |
| 83 | 10000 | 0.2 | 954645 | 69.0 | 958865 | 116.0 | -4220 | 1.7 |
| 84 | 10000 | 0.2 | 3537453 | 148.8 | 3529326 | 128.0 | 8127 | 0.9 |
| 85 | 10000 | 0.2 | 299702 | 172.0 | 299427 | 124.0 | 275 | 0.7 |
| 86 | 10000 | 0.2 | 4652467 | 106.7 | 4653917 | 116.0 | -1450 | 1.1 |
| 87 | 10000 | 0.2 | 542995 | 88.8 | 542583 | 115.0 | 412 | 1.3 |
| 88 | 10000 | 0.2 | 2430925 | 101.6 | 2426515 | 107.0 | 4410 | 1.1 |
| 89 | 10000 | 0.2 | 4664901 | 106.7 | 4653846 | 88.0 | 11055 | 0.8 |
| 90 | 10000 | 0.2 | 296604 | 153.8 | 296251 | 119.0 | 353 | 0.8 |
| 91 | 10000 | 0.2 | 2542752 | 99.8 | 2545270 | 114.0 | -2518 | 1.1 |
| 92 | 10000 | 0.2 | 547389 | 116.9 | 546725 | 119.0 | 664 | 1.0 |
| 93 | 10000 | 0.2 | 8695308 | 116.6 | 8661439 | 63.0 | 33869 | 0.5 |
| 94 | 10000 | 0.2 | 426909 | 126.1 | 426758 | 125.0 | 151 | 1.0 |
| 95 | 10000 | 0.2 | 3533974 | 149.8 | 3508360 | 118.0 | 25614 | 0.8 |
| 96 | 10000 | 0.2 | 961109 | 60.0 | 959744 | 1.8 | 1365 | 0.0 |

B1 (continued).  PR2 comparison of Objective Value and Time using the 96 Test Problems